# Loss-analysis via Attention-scale for Physiologic Time Series

Jiawei Yang and Jeffrey M. Hausdorff*

Physiologic signals have properties across multiple spatial and temporal scales, which can be shown by the *complexity-analysis* of the coarse-grained physiologic signals by scaling techniques such as the *multiscale*. Unfortunately, the results obtained from the coarse-grained signals by the multiscale may not fully reflect the properties of the original signals because there is a loss caused by scaling techniques and the same scaling technique may bring different losses to different signals. Another problem is that multiscale does not consider the key observations inherent in the signal. Here, we show a new analysis method for time series called the *loss-analysis* via *attention-scale*. We show that multiscale is a special case of attention-scale. The loss-analysis can complement to the complexity-analysis to capture aspects of the signals that are not captured using previously developed measures. This can be used to study ageing, diseases, and other physiologic phenomenon. We find that heart rate variability and gait signals have recognizable losses in scaling under the attention-scale mechanism.

To study the physiologic systems, there is a great interest in the *complexity-analysis* [1-4] of the physiologic signals obtained from the physiologic systems by recording physiologic events such as the heart beating or a muscle contracting. Typically, entropy-based [5-7] methods are employed to quantify the complexity of physiologic signals to uncover hidden patterns of signals. However, they cannot yield important insights for signals' properties over different scales [8, 9].

Therefore, the complexity-analysis of the coarse-grained physiologic signals by scaling techniques has been the focus of considerable attention [8 -10]. For example, a typical scaling technique called *multiscale* [8] is used to scale the interbeat time series [8, 9] and the gait time series [10] after which the complexity analysis of the coarse-grained time series was performed by entropy-based methods. For example, in [8], after quantifying the complexity of the coarse-grained interbeat time series, it found the coarse-grained interbeat time series from healthy young subjects always has higher complexity than that from healthy elderly subjects. The result is consistent with the hypothesis of loss of complexity with age [11].

However, the analysis method to analyze physiologic signals by analyzing their coarse-grained versions may not fully reflect the property of the original physiologic signals. This is because there is a loss between the original physiologic signals and their coarse-grained versions, caused by the scaling techniques used. We refer this loss to *scaling-loss*.

Thus, to complement the complexity-analysis of the original physiologic signals or their coarse-grained versions, we propose another analysis method called *loss-analysis*, which measures the scaling-loss between the physiologic signals and their coarse-grained versions by scaling techniques. We can see the loss-analysis has two important cores: the measure of the scaling-loss and the scaling technique utilized.

To measure the scaling-loss, we introduce two forms of loss: *complexity-loss* and *similarity-loss*. We can use the *difference* of the complexity between original and the coarse-grained signals quantified independently by any entropy-methods as complexity-loss. We can use the *distance* between original and coarse-grained signals as the similarity-loss. We can see that the main difference between complexity-loss and similarity-loss is whether their calculations involve the original and coarse-grained signals independently or dependently.

Scaling technique is another important core of the loss-analysis. A typical scaling technique is multiscale [8]. Given a time series of observations, multiscale [2] works in three steps: (1) select $(i\tau)^{th}$ observation, where $\tau$, an integer, is the *scale factor*, $i$ is an integer, and use them as the *segment observations*, (2) segment the time series into non-overlapping sub-series with the locations of the segment observations, and (3) calculate the means of sub-series as the coarse-grained time series. Formally, given a one-dimensional discrete time series of observations, $X = \{x_1, \ldots, x_N\}$, multiscale [8] scales the series $X$ to a construct consecutive coarse-grained time series, $Y = \{y^{(\tau)}\}$, determined by the scale factor, $\tau$, according to the equation: $y_i^{(\tau)} = 1/\tau \sum_{j=(i-1)\tau+1}^{i\tau} x_j, 1 \leq i \leq N/\tau$. Multiscale has been widely cited in basic research and a variety of applications due to its simplicity but high efficiency [12]. However, we can see that the segment observations play a very vital role in scaling, but multiscale [8] selects them without considering any information about the time series such as length, distribution, or structure of time series. This may become problematic when there are important structures or key observations inherent in time series, which may not be emphasized by merely tuning the scale factor.

Here, we propose the *attention-scale* technique, which pays attention to key observations inherent in the physiologic signals. Multiscale [8] is a special case of attention-scale. Their differences are the key observations which they pay attention to.

## Results

In this paper, we used Shannon entropy [5] difference as complexity-loss and the Kolmogorov-Smirnov (KS) statics [13] as similarity-loss. According to the key observations used in this paper, we category the attention-scale techniques and name them as multiscale (MS) [8], peak attention-scale (PAS), occurrence attention-scale (OAS), and median attention-scale (MAS). We also performed the complexity-analysis of the coarse-grained signals with Shannon entropy to compare to the results of the loss-analysis. The Shannon entropy results of the coarse-grained signals scaled by MS [2] with the scale factor of one correspond to the Shannon entropy of the original signals. We tested simulated white and 1/f noise time series, real-world gait time series, and real-world interbeat intervals time series downloaded from PhysioNet [14].

**Simulated white and 1/f noise time series**. From Fig. 1, we can see that (1) except PAS, only the similarity-loss-analysis can distinguish 1/f and white noise. This shows the advantage of similarity-loss-analysis. It also shows that white noise has more similarity-loss than 1/f noise no matter which scaling is used. This shows white noise is more sensitive to all the attention observations used. We can see (2) that only the PAS can separate the noise no matter which analysis is performed, and this shows 1/f and white signals are more sensitive to the attention observations defined by PAS. This indicates that different signals can show different properties in scaling using different attention observations. We can also see (3) that 1/f noise has more complexity-loss than white noise. This indicates the complexity-loss and similarity-loss are two independent and uncorrelated losses.

**Gait time series with walking states**. From Fig. 2, we can see that (1) the complexity-analysis of coarse-grained time series by MS [8] and PAS and the similarity-loss-analysis with MS, PAS, and OAS can separate the groups between unconstrained and metronomic walking states. However, only the complexity-loss-analysis with OAS can separate the groups between slow speed and other speed. This indicates different walking states are sensitive to different key observations used and loss-analysis can uncover more insight which is a vital complementary to complexity-analysis. We also see that (2) the complexity-analysis of coarse-grained time series only with PAS can statistically significantly separate FAST, NORM, and SLOW ($p$-values < 0.01), this shows PAS can capture the speed better than other scaling methods including MS [8].

**Gait time series with aging and diseases**. From Fig. 3, we can see that *Parkinson* is sensitive to all the key observations and statistically significantly separable form *Young* and *Elderly* over most scale factors ($p$-values < 0.01). We can also see that (2) except the results from the similarity-loss-analysis with MAS, the other results are consistent. This indicates again that similarity-loss and complexity-loss are uncorrected.

**Gait time series with maturation**. From Fig. 4, we can see that (1) from the complexity-analysis of the coarse-grained signals by OAS and MAS and the complexity-loss-analysis with OAS and MAS, it shows the complexity-analysis is uncorrected to complexity-loss-analysis. We can also see that (2) with the complexity-loss-analysis, no matter which scaling is used, the three groups can be statistically significantly separated same time ($p$-values < 0.01). This shows the robustness of complexity-loss-analysis.

From Table I, we can see that PAS has better correlation with *age* [15] than all other entropy-based measures tested. As the combination results of PAS, MS and *height* [15] can reach the most correlation to *age* [15], this indicates that PAS is an independent measure which can provide new information that is not provided by previously developed measures.

**Heart rate variability with aging and diseases**. From Fig. 5, we can see that (1) only PAS with the complexity-analysis of the coarse-grained signals and complexity-loss-analysis can statistically significantly separate all groups same time ($p$-values < 0.01 e.g. scale factor =1). This shows the interbeat intervals are more sensitive to the key observation used by PAS. We can also see (2) that it is not always healthy subjects have most complexity values or complexity-loss values. This tells that when trying to conclude which group of subjects have most complexity or complexity-loss, the prior of using which key observations must be given. (Therefore, the conclusions in [8] hold only when key observations used as defined by MS [8].)

## Discussion

**The complexity-analysis and loss-analysis**. The complexity-analysis of original physiologic signals cannot uncover the properties of physiologic signals over different scales, while the complexity-analysis of coarse-grained physiologic signals by scaling techniques can reveal to the properties of physiologic signals over different scales but cannot fully reflect the properties of original physiologic signals due to the loss caused by scaling techniques. The loss-analysis can bring to light the physiologic signals' loss over different scales can complement to existing analysis methods.

**Why does attention-scale work**? A physiologic system has complex functions. The system stays at different physiologic states when expressing different functions. The important states associated with important functions are the key to keep system functionally running. However, the system's adaption to the interference from the environment such as stress will produce noisy states, which do not necessary for the system's functions. The appearance of noisy states will affect the occurrences of the important states.

Attention-scale is a technique to average the important states with the noisy states around them. The important states will change when averaging with the noisy states, and the change can be seen as the system's adaption ability to the environment. Different systems with different physiologic conditions such as ageing or diseases should have different adaption ability to the environment. Given a series of observation recorded from a physiologic system in the order of time, we can see each observation as one state of the system, the loss between the original observations and coarse-grained observations by scaling techniques can be used to monitor the physiologic system's adaption ability to the environment.

**Physiologic signals' properties over different scales**. Attention-scale brings new insights to physiologic signals' properties over different scales: when the signals are coarse-grained by filtering the noisy observations around different key observations by attention-scale, different physiologic signals from different physiologic phenomenon have different degree of change. This change can be potential for solving or detecting problems in a wide range of domains.

**Applicable to industry and clinical research**. The physiologic signals such as heart rate variability and gait time series have been widely monitored in hospitals, clinical centers, or by smart devices and wearable devices such as smartphones, smartwatches, and smart bracelets which provide affordable ways for normal users to monitor their physiological condition by themselves. However, the tool to study the long-term physiologic signals is rare.

Attention-scale may be a potential technique with the ability to convert the long-term signals to short signals. Attention-scale also provides a way to explore the signals' property over different scales. The loss-analysis via attention-scale is also an important complementary to the analysis of the physiologic signals.

**Limitations**. Attention-scale has to pre-define the key observation which benefits to the analysis. Different physiologic signals recorded from different physiologic events or from different physiologic systems may require different attention observations. Even four kinds of key observations have been shown in this paper, more other key observations may be needed to uncover more physiological phenomenon.

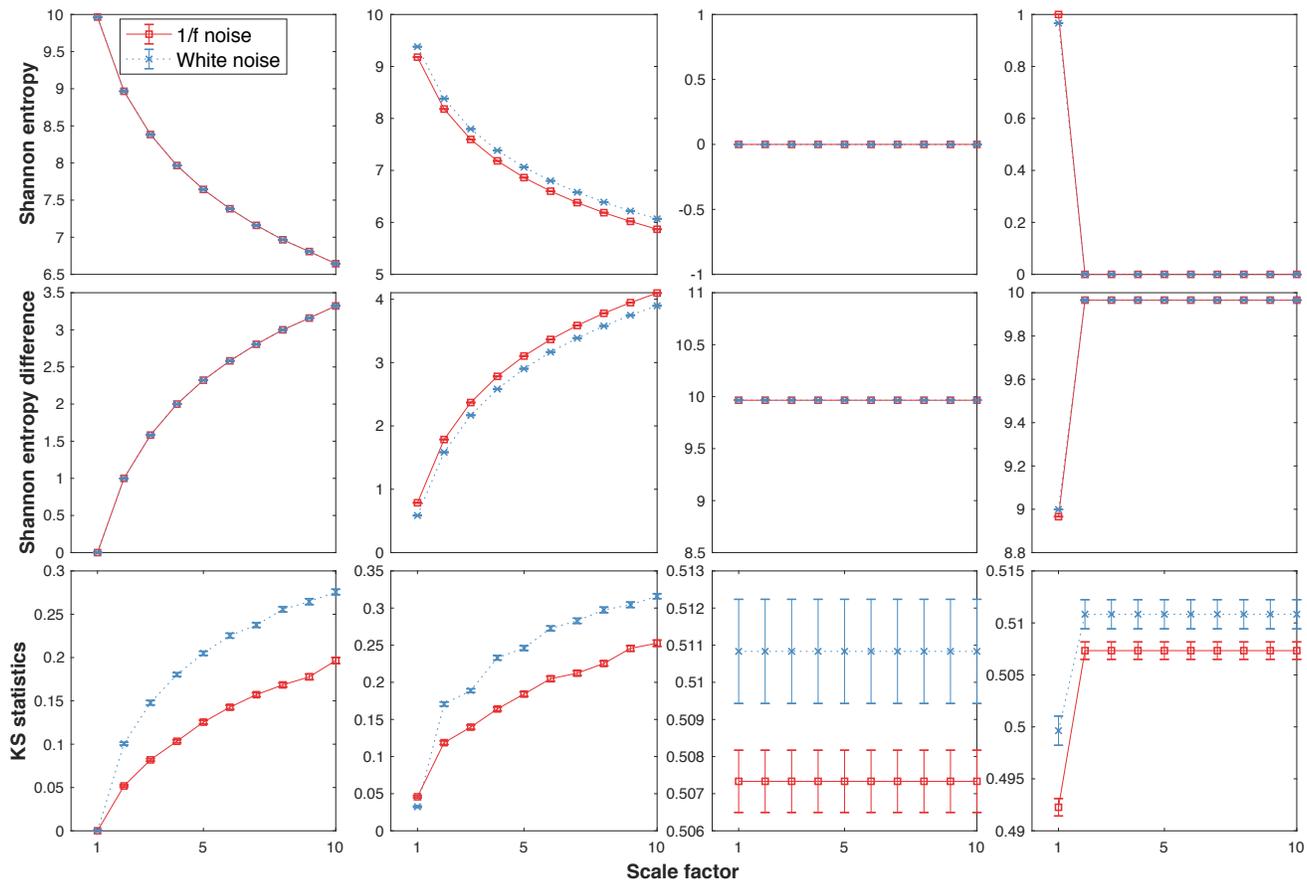

Fig. 1 Scaling analysis of 30 simulated Gaussian distributed (mean zero, variance one) white and 1/$f$ noise time series. Symbols represent the mean values of entropy, and bars represent the standard error (SE = standard deviation / $\sqrt{n}$, where $n$ is the number of subjects). The first, second, third and fourth column show the signals scaled by the MS [8], the PAS, the FAS, and the MAS respectively, while the first, second, third and fourth row show the results of the complexity-analysis of coarse-grained signals (Shannon entropy), the complexity-loss-analysis (Shannon entropy difference between the original signal and its coarse-grained version), and the similarity-loss-analysis (KS statistic).

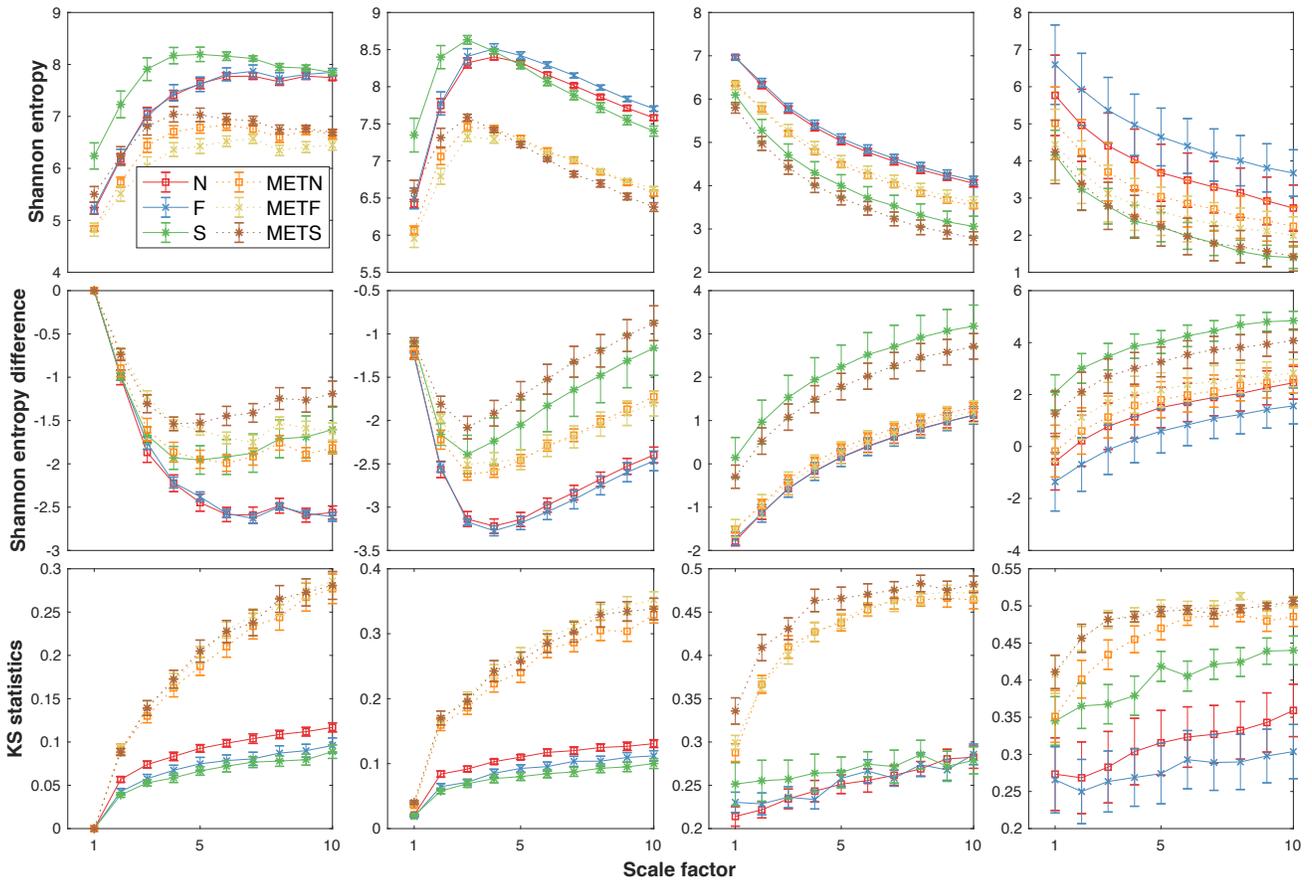

Fig. 2 Scaling analysis of gait time series derived from unconstrained and metronomic walking from 10 young and healthy subjects. Symbols represent the mean values of entropy, and bars represent the standard error (SE =standard deviation / $\sqrt{n}$, where $n$ is the number of subjects). The first, second, third and fourth column show the signals scaled by the MS [8], the PAS, the FAS, and the MAS respectively, while the first, second, third and fourth row show the results of the complexity-analysis of coarse-grained signals (Shannon entropy), the complexity-loss-analysis (Shannon entropy difference between the original signal and its coarse-grained version), and the similarity-loss-analysis (KS statistic).

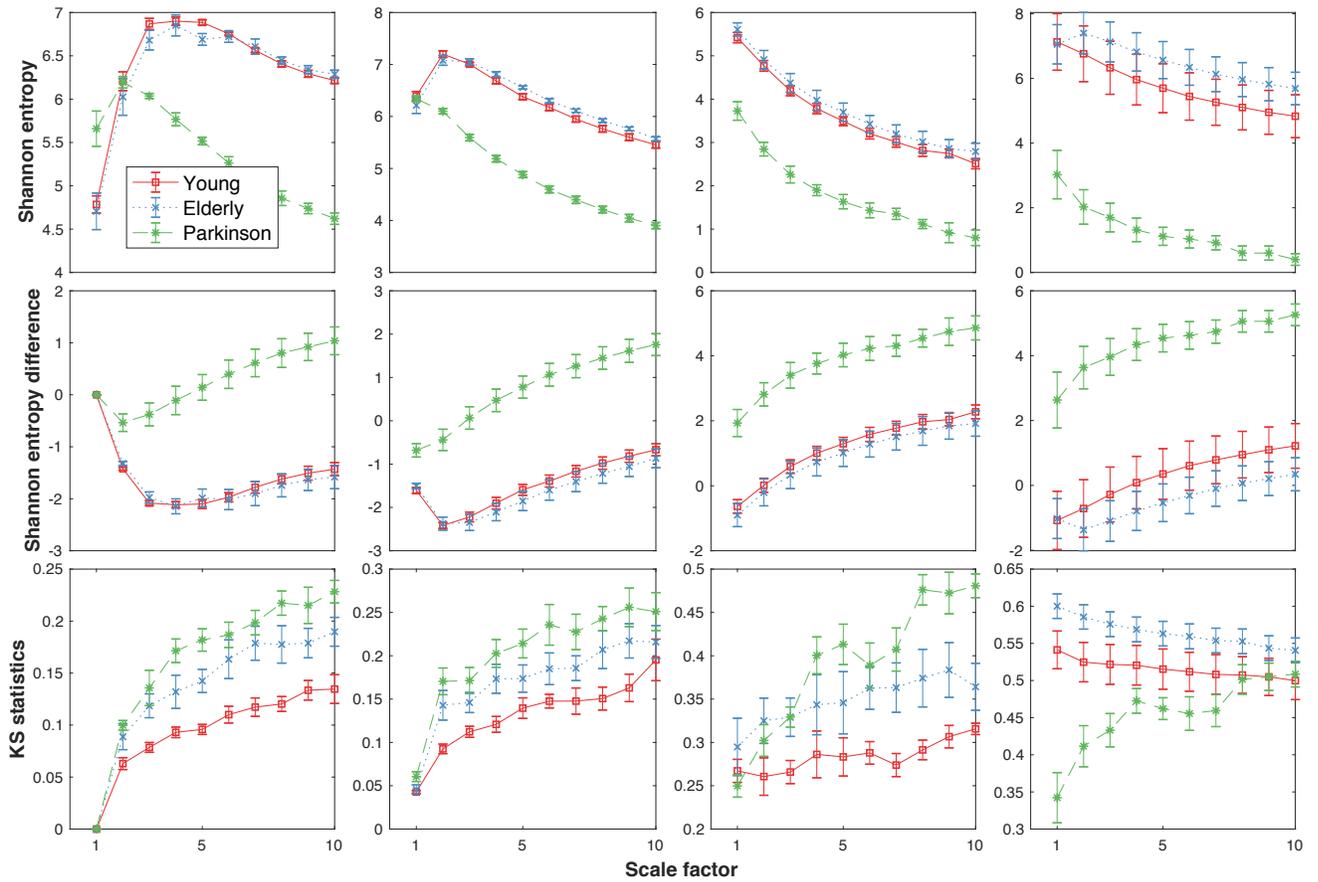

Fig. 3 Scaling analysis of gait time series derived from 5 healthy young adults (23 - 29 years old), 5 healthy old adults (71 - 77 years old), and 5 older adults (60 - 77 years old) with Parkinson's disease. Symbols represent the mean values of entropy, and bars represent the standard error (SE =standard deviation / $\sqrt{n}$, where $n$ is the number of subjects. The first, second, third and fourth column show the signals scaled by the MS [8], the PAS, the FAS, and the MAS respectively, while the first, second, third and fourth row show the results of the complexity-analysis of coarse-grained signals (Shannon entropy), the complexity-loss-analysis (Shannon entropy difference between the original signal and its coarse-grained version), and the similarity-loss-analysis (KS statistic).

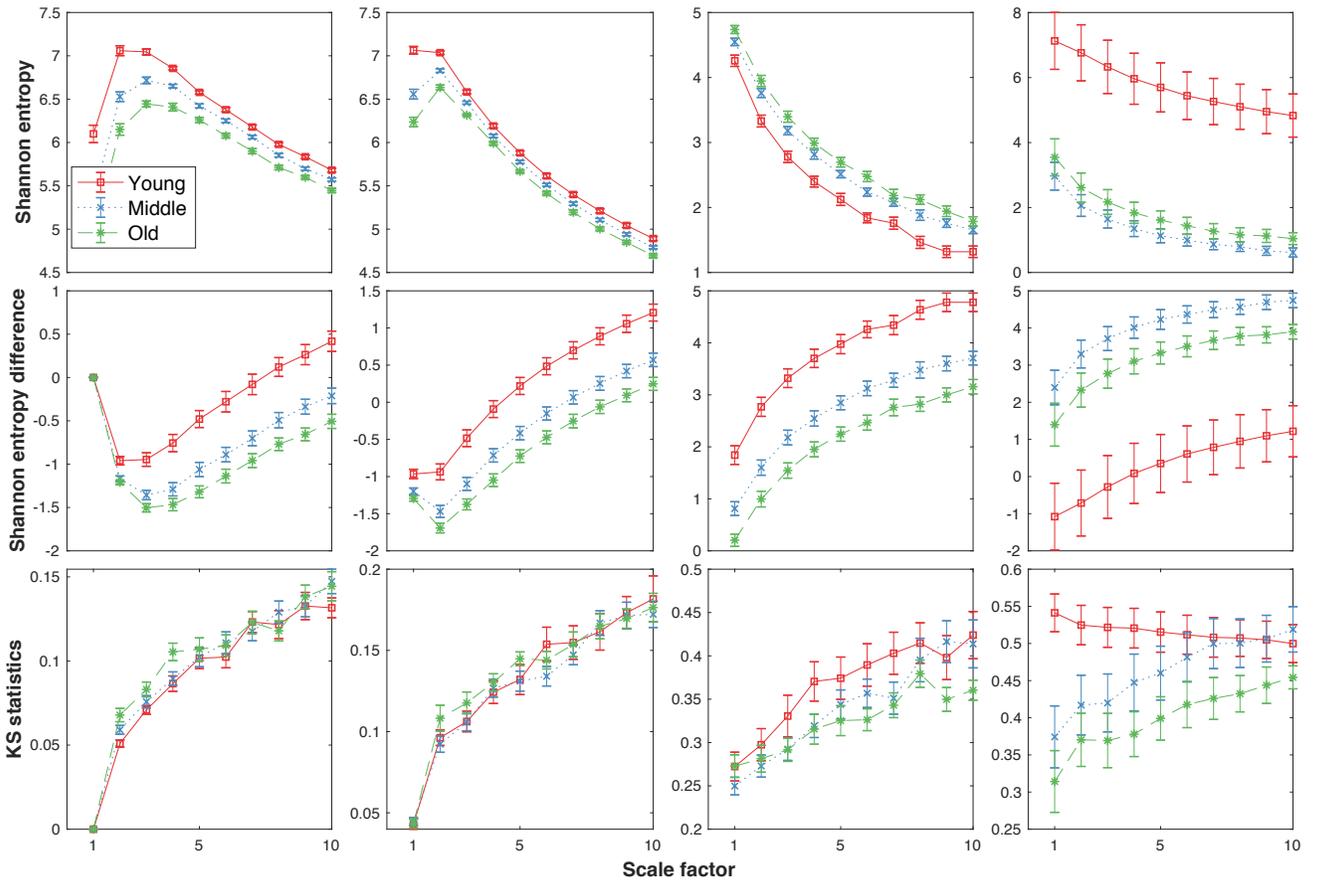

Fig. 4 Scaling analysis of gait time series derived from 43 healthy children ranging in age from 40 months to 163 months (11 young children aging between 40 and 58 months, 20 middle young children aging between 79 and 94 months, 12 old children aging between 133 and 163 months). Symbols represent the mean values of entropy, and bars represent the standard error (SE = standard deviation / $\sqrt{n}$, where $n$ is the number of subjects). The first, second, third and fourth column show the signals scaled by the MS [8], the PAS, the FAS, and the MAS respectively, while the first, second, third and fourth row show the results of the complexity-analysis of coarse-grained signals (Shannon entropy), the complexity-loss-analysis (Shannon entropy difference between the original signal and its coarse-grained version), and the similarity-loss-analysis (KS statistic).

Table I: Pearson correlation and Welch's t-test for binary groups. [26]

| | | | | | | Correlation | Welch's t-test (*p*-value) | | |
|---|---|---|---|---|---|---|---|---|---|
| | | | | | | Age | Young, Middle | Young, Old | Middle, Old |
| Other measurements | | | | | Height [15] | 0.957 | 3.73E-10 | 4.10E-11 | 8.53E-08 |
| | | | | | Weight [16] | 0.902 | 1.85E-07 | 3.77E-07 | 1.31E-05 |
| | | | | | Leg length [17] | 0.898 | 3.79E-08 | 2.79E-09 | 2.74E-06 |
| | | | | | Speed [15] | 0.653 | 1.01E-05 | 2.00E-07 | 4.20E-02 |
| Gait time series | Statistics | | | | Mean [16] | 0.753 | 7.65E-03 | 7.88E-07 | 1.29E-04 |
| | | | | | RMSSD [16] | 0.385 | 4.63E-02 | 1.44E-06 | 1.94E-01 |
| | | | | | SDNN [16] | -0.200 | 3.40E-01 | 2.52E-04 | 6.52E-02 |
| | | | | | NN50 [16] | -0.808 | 1.52E-08 | 3.67E-11 | 5.06E-05 |
| | | | | | PNN50 [16] | -0.778 | 3.62E-07 | 6.10E-10 | 2.53E-04 |
| | Entropy | No scaling | | | Shannon_ent [5] | -0.739 | 1.62E-05 | 3.81E-08 | 8.34E-04 |
| | | | | | Renyi_ent [6] | -0.728 | 2.71E-05 | 8.96E-08 | 9.42E-04 |
| | | | | | Tsallis_ent [7] | -0.729 | 3.26E-06 | 8.22E-08 | 1.81E-03 |
| | | | | | Attention_ent [17] | -0.267 | 6.28E-01 | 8.82E-02 | 1.11E-01 |
| | | | | | Spectral_ent [18] | -0.380 | 1.23E-01 | 2.97E-02 | 3.31E-01 |
| | | | | | SVD_ent [19] | -0.513 | 6.81E-01 | 1.23E-04 | 2.14E-03 |
| | | | | | GHV_ent [20] | -0.376 | 5.66E-01 | 1.09E-01 | 1.70E-01 |
| | | | | | Permutation_ent [21] | -0.743 | 1.54E-03 | 7.68E-07 | 4.27E-04 |
| | | | | | Edge_perm_ent [22] | 0.225 | 5.84E-01 | 7.01E-02 | 1.38E-01 |
| | | | | | Bubble_ent [23] | -0.126 | 5.18E-02 | 2.69E-01 | 4.25E-01 |
| | | | | | Approxim_ent [24] | 0.116 | 1.73E-02 | 7.12E-01 | 2.82E-02 |
| | | | | | Sample_ent [25] | 0.320 | 2.08E-01 | 1.08E-01 | 2.00E-02 |
| | | With scaling | MSE [8] | | | 0.395 | 6.19E-01 | 1.85E-02 | 1.36E-02 |
| | | | MS [8] | Fixed scale factor | Shannon entropy [5] | -0.830 | 8.34E-08 | 3.37E-07 | 2.61E-04 |
| | | | PAS (proposed) | | | -0.833 | 2.31E-07 | 1.16E-09 | 2.95E-05 |
| | | | FAS (proposed) | | | 0.586 | 7.70E-04 | 1.52E-05 | 3.02E-02 |
| | | | MAS (proposed) | | | 0.554 | 1.65E-02 | 1.21E-04 | 1.63E-02 |
| | | | MS [15] | Dynamic scale factor | | -0.930 | 1.20E-06 | 1.93E-11 | 1.76E-07 |
| | | | PAS (proposed) | | | -0.934 | 2.32E-06 | 4.82E-11 | **1.17E-10** |
| | | | FAS (proposed) | | | -0.307 | 9.99E-01 | 2.22E-01 | 1.39E-01 |
| | | | MAS (proposed) | | | 0.194 | 5.39E-01 | 2.31E-01 | 4.24E-01 |
| | | Combination of height, MS, and PAS | | | | **-0.970** | **1.50E-15** | **7.51E-13** | 1.50E-10 |

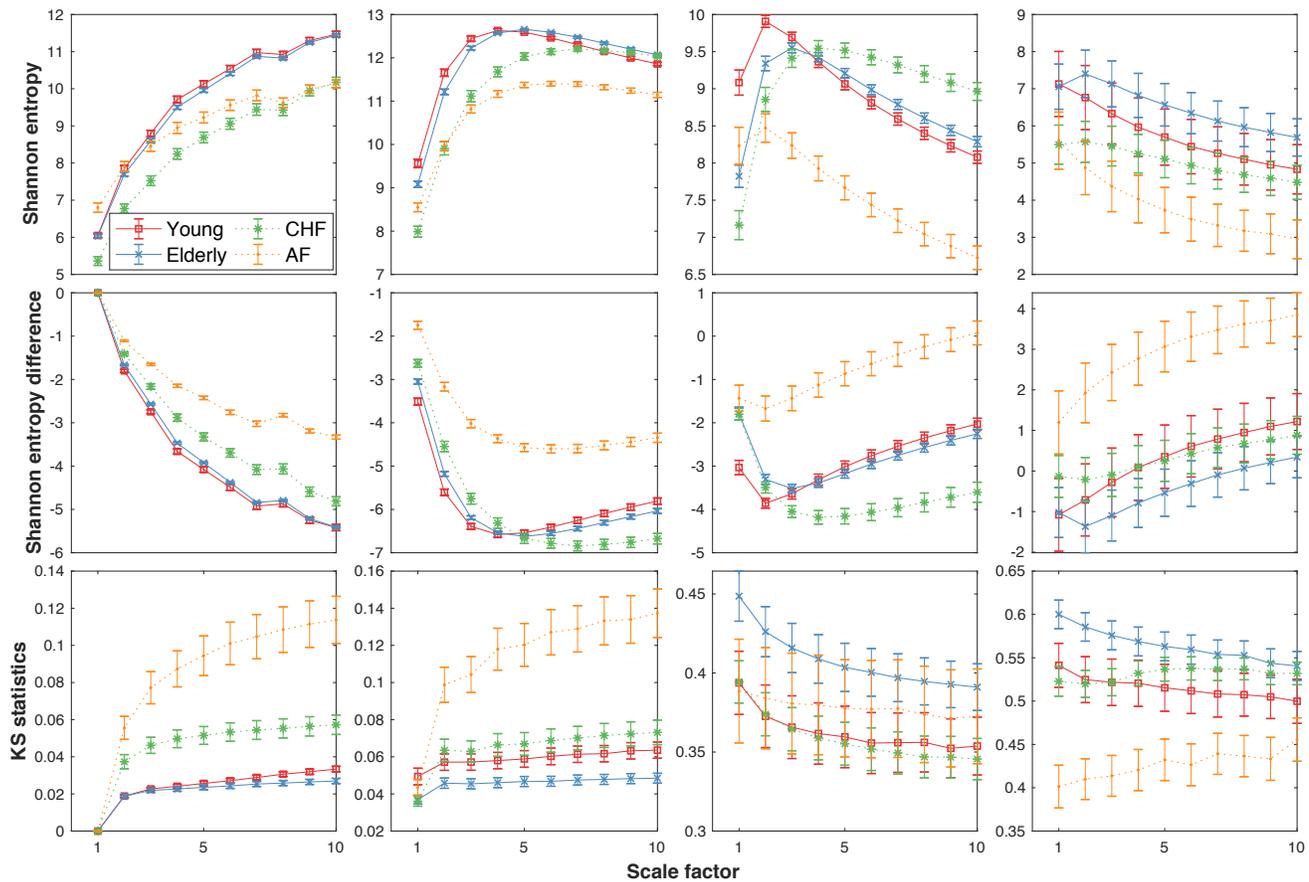

Fig. 5 Scaling analysis of interbeat intervals time series derived from healthy subjects with age ≤55 (young), healthy subjects with age >55 (elderly), subjects with congestive heart failure (CHF), and subjects with atrial fibrillation (AF). Symbols represent the mean values of entropies, and bars represent the standard error (SE =standard deviation / $\sqrt{n}$, where n is the number of subjects). The first, second, third and fourth column show the signals scaled by the MS [8], the PAS, the FAS, and the MAS respectively, while the first, second, third and fourth row show the results of the complexity-analysis of coarse-grained signals (Shannon entropy), the complexity-loss-analysis (Shannon entropy difference between the original signal and its coarse-grained version), and the similarity-loss-analysis (KS statistic).

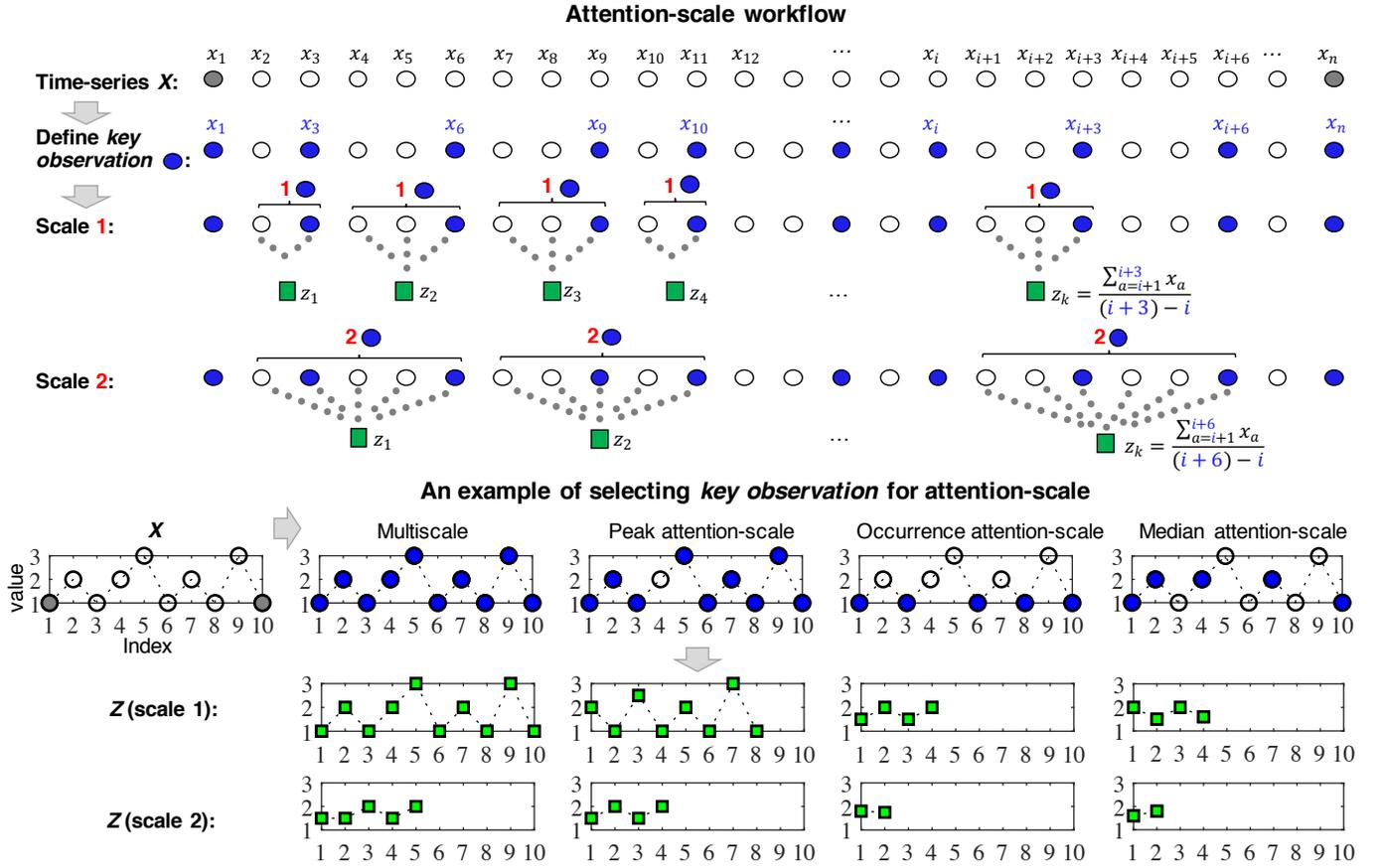

Fig. 6), (top) Schematic illustration of the attention-scale procedure. (bottom) An example of selecting key observations: MS [8] uses all observations; PAS uses observation that are bigger or smaller than the observations before and after it same time; OAS uses the observation that has most occurrence; and MAS uses the observation which is the median of the observations.

## Outlook

We notice that the proposed loss-analysis via attention-scale is potentially applicable for solving a wide range of problems in different domains. In future research, we aim to explore more key observations and more loss measure between original signals and its coarse-grained versions.

## Methods

**Attention-scale**. Here, we propose an *attention-scale* technique, which pays attention to key observations, called *attention observations*, inherent in time series of signals. Given a one-dimensional discrete time series of observations, $X = \{x_1, \ldots, x_N\}$, attention-scale works in four steps: (1) define attention observations inherent in time series $X$ which we want to pay attention to, (2) select the $(i\tau)^{th}$ attention observation, where $\tau$, an integer, is the *scale factor*, and $i$ is an integer $1 \leq i \leq N/\tau$; then use them as the segment observations, (3) segment the $X$ into non-overlapping sub-series by the locations of the segment observations; (4) calculate the means of sub-series as the coarse-grained time series. Formally, attention-scale scales the series $X$ to a construct consecutive coarse-grained time series, $Z = \{z^{(\tau)}\}$, determined by the scale factor, $\tau$, according to the equation: $z_i^{(\tau)} = 1/t \sum_{j=1+I_{(i-1)\tau}}^{I_{i\tau}} x_j$, where $1 \leq i \leq N/\tau$, $t = (I_{i\tau} - I_{(i-1)\tau})$, and $I_{i\tau}$ is the index of the $(i\tau)^{th}$ attention observation, which means that $x_{I_{i\tau}}$ is the attention observation as pre-defined. The workflow of attention-scale is illustrated in Fig. 6 (top).

**Attention observations.** In this paper, we consider three types of attention observations:
(1) the observation $x_k$ which satisfies $(x_k - x_{k-1})(x_k - x_{k+1}) > 0$;
(2) the observation $x_k$ which has the most occurrence in $X$;
(3) the observation $x_k$ which is the median of all different observations in $X$.

The first type of attention observation considers the local relationship of the observations, and it does not require knowing the information of the *whole* signals. Since this type of attention observations look like *peaks* in the time series, we refer the attention-scale using the first type of attention observation as peak attention-scale (PAS). The second type of attention observation considers the global relationship of the observations in the view of observations' occurrences. We refer the attention-scale using the second type of attention observation as occurrence attention-scale (OAS). The third type of attention observation considers the global relationship of the observations in the view of observations' values. We refer the attention-scale using the third type of attention observation as median attention-scale (MAS). An example of these three key observation is illustrated in Fig. 6 (bottom).

If seeing each observation in time series as one state of a physiologic system of adjustment to the change of environment, the MS [8] and the PAS reflect the property of local adjustment, while the FAS and the MAS reflect the property of global adjustment. MS [8] measures how evenly all states of the system are outcome. PAS measures how the states referring to sudden changes are outcome. FAS measures how the states in the majority are outcome. MAS measures how the states on average (median) are outcome. Since MS [8] and PAS do not need to know the information of the whole signals, they can be applied to some applications to monitor the system real-time.

**Multiscale (MS) and Attention-scale**. MS [8] is a special case of attention-scale when defining *all* observations in signal as attention observations. In the view of attention, MS [8] pays attentions equally to all observations in signals. It has the advantage that it can measure how equally the observations are important. However, it also has the disadvantage that it is too greedy to pay attention to all observations in the time series because different observations may play different roles when reflecting the functions of physiologic systems and some of the observations may be just noises, which do not carry any useful information.

**Loss-analysis via Attention-scale (LAAS)**. LAAS is to measure the change between original signals and its coarse-grained versions by attention-scale technique. We call this change as *loss* and the analysis of this change as *loss-analysis* via attention-scale. How to measure the loss is an open problem. Here, we introduce two forms of measures. One form is to quantify the original signals and its coarse-grained versions independently and then use their difference as loss, called *complexity-loss*. Another form is to calculate the distance between this two time-series and use the distance as loss, called *similarity-loss*.

**Complexity-loss-analysis.** Complexity-loss-analysis treats the original signals and its coarse-grained versions as independent signals, quantifies their complexity respectively, and study their difference in complexity caused by scaling. The methods shown in [1-4], [5-8], and [16-25] can be used to quantify the complexity of signals.

**Similarity-loss-analysis.** Similarity-loss-analysis treats the original signals and its coarse-grained versions as dependent signals and studies their distance caused by scaling. The methods shown in [27] such as dynamic time warping distance, Frechet distance, or correlation coefficient can be used to calculate the distance between time series.

# Data availability
The data is available at PhysioNet [14].

# Code availability
The code is available at github.com and PhysioToolkit [14].

# References (will work on reference at last)

[1] S. M. Pincus, Ann. N.Y. Acad. Sci. 954, 245 (2001), and references therein.
[2] A. Porta, S. Guzzetti, N. Montano, R. Furlan, M. Pagani, A. Malliani, and S. Cerutti, IEEE Trans. Biomed. Eng. 48, 1282 (2001); M. Palus, Physica (Amsterdam) 93D, 64 (1996); N. Wessel, A. Schumann, A. Schirdewan, A. Voss, and J. Kurths, Lect. Notes Comput. Sci. 1933, 78 (2000).
[3] A. L. Goldberger, C.-K. Peng, and L. A. Lipsitz, Neurobiol. Aging 23, 23 (2002).
[4] J. S. Richman and J. R. Moorman, Am. J. Physiol. 278, H2039 (2000).
[5] C. E. Shannon, "A mathematical theory of communication", The Bell System Technical Journal, vol. 27, no. 3, pp. 379 - 423, 1948.
[6] A. Rényi, "On measures of entropy and information", in Proceedings of the Fourth Berkeley Symposium on Mathematical Statistics and Probability, Volume 1: Contributions to the Theory of Statistics, 1961.
[7] C. Tsallis, "Possible generalization of boltzmann-gibbs statistics", Journal of Statistical Physics, vol. 52, no. 1-2, pp. 479-487, 1988.
[8] M. Costa, A. L. Goldberger, and C.-K. Peng, Multiscale entropy analysis of complex physiologic time series, Phys. Rev. Lett., vol. 89, no. 6, p. 68102, 2002.
[9] M. Costa, A. L. Goldberger, and C.-K. Peng, Multiscale entropy analysis of biological signals, Phys. Rev. E, vol. 71, no. 2, p. 21906, 2005.
[10] M. Costa, C. K. Peng, A. L. Goldberger, and J. M. Haus- dorff, Multiscale entropy analysis of human gait dynamics, Physica A: Statistical Mechanics and its Applications 330, 53, 2003.
[11] A. Goldberger, C. Peng and L. Lipsitz, What is physiologic complexity and how does it change with aging and disease?, Neurobiol. aging, vol. 23, no. 1, pp. 23-26, 2002.
[12] D. R. Chialvo, Physiology: Unhealthy surprises, Nature, vol. 419, no. 6904, pp. 263, Sep. 2002
[13] Hodges, J.L. Jr., The Significance Probability of the Smirnov Two-Sample Test, Arkiv fiur Matematik, 3, No. 43, 469-86, 1958.
[14] A. L. Goldberger, L. A. N. Amaral, L. Glass, J. M. Hausdorff, P. C. Ivanov, R. G. Mark, et al., "Physiobank physiotoolkit and physionet: components of a new research resource for complex physiologic signals", Circulation, vol. 101, no. 23, pp. e215-e220, 2000.
[15] Hausdorff et al., Journal of Applied Physiology, vol. 86: 1040-1047, 1999.
[16] L. Wang, W. Zhou, Q. Chang, J. Chen and X. Zhou, "Deep Ensemble Detection of Congestive Heart Failure Using Short-Term RR Intervals", in IEEE Access, vol. 7, pp. 69559-69574, 2019.
[17] Jiawei YANG.. Attention entropy
[18] Inouye, K. Shinosaki, et al., "Quantification of EEG irregularity by use of the entropy of the power spectrum", Electroencephalogr. Clin. Neurophysiol., vol. 79, no. 3, pp. 204–210, 1991.
[19] O. Alter, P. O. Brown, and D. Botstein, "Singular value decomposition for genome-wide expression data processing and modeling", Proc. Natl. Acad. Sci., vol. 97, no. 18, pp. 10101–10106, 2000.
[20] G. I. Choudhary, W. Aziz, I. R. Khan, S. Rahardja, and P. Fränti, "Analysing the Dynamics of Interbeat Interval Time Series Using Grouped Horizontal Visibility Graph", IEEE Access, vol. 7, pp. 9926–9934, 2019.
[21] C. Bandt and B. Pompe, "Permutation entropy: a natural complexity measure for time series", Phys. Rev. Lett., vol. 88, no. 17, p. 174102, 2002.
[22] Z. Huo, Y. Zhang, L. Shu, and X. Liao, "Edge Permutation Entropy: An Improved Entropy Measure for Time-Series Analysis", in IECON 2019 - 45th Annual Conference of the IEEE Industrial Electronics Society, vol. 1, pp. 5998–6003, 2019.
[23] G. Manis, M. D. Aktaruzzaman, and R. Sassi, "Bubble entropy: an entropy almost free of parameters", IEEE Trans. Biomed. Eng., vol. 64, no. 11, pp. 2711–2718, 2017.
[24] S. M. Pincus, "Approximate entropy as a measure of system complexity", Proc. Natl. Acad. Sci., vol. 88, no. 6, pp. 2297–2301, 1991.
[25] J. S. Richman and J. R. Moorman, "Physiological time-series analysis using approximate entropy and sample entropy", Am. J. Physiol. Circ. Physiol., vol. 278, no. 6, pp. H2039--H2049, 2000.
[26] When calcuating MSE or attention-scale with fixed scal factor, the scale factor that provides best results varying from 1 to 15 is used. When calcuating attention-scale with dynamic scal factor, the scale factor is set to (height -1)/18. When combing the height, MS, and PAS, the sum of Shannon entropy values of coarse-grained time series by PAS with scale facor of (height -1)/18 and by applying twice MS with scale facor of (height -1)/9 is used as final entropy value.
[27] Tak-chung Fu, "A review on time series data mining", Engineering Applications of Artificial Intelligence, 24(1), pp. 164-181,2011,.

# Acknowledgements

~~we have no one to thanks so far.~~

# Author Contributions